\documentclass[runningheads]{llncs}
\usepackage[T1]{fontenc}

\usepackage{graphicx}
\usepackage{float}
\usepackage{multirow}
\usepackage{tabularx}
\usepackage{xcolor}

\newcommand{\tocheck}[1]{#1}
\newcommand{\tohide}[1]{}

\usepackage{enumitem} 
\usepackage{amsmath}
\usepackage{amssymb}
\usepackage{oz}
\usepackage{arydshln} 
\usepackage{tabularray}

\usepackage{threeparttable} 
\newcommand*{\semanticTripleFont}{\fontfamily{qcs}\selectfont}
\newcommand*{\ssubject}{\text{{\semanticTripleFont{subject}}}}
\newcommand*{\spredicate}{\text{{\semanticTripleFont{predicate}}}}
\newcommand*{\sobject}{\text{{\semanticTripleFont{object}}}}
\newcommand{\abs}[1]{\,\mid\!\!#1\!\!\mid\,}

\usepackage{xspace}
\newcommand{\myparagraph}[1]{\vspace{0.5em}\noindent\textbf{#1}\xspace}

\newcommand{\ch}[1]{#1}

\usepackage{orcidlink}
\usepackage{academicons}
\begin{document}
\title{
HOEG: A New Approach for Object-Centric Predictive Process Monitoring
}

\author{Tim K. Smit\inst{1}\orcidlink{0009-0000-4743-3522} 
\and
Hajo A. Reijers\inst{1}\orcidlink{0000-0001-9634-5852} \and
Xixi Lu\inst{1}\orcidlink{0000-0002-9844-3330} 
}
\authorrunning{Smit et al.}

\institute{Utrecht University, Utrecht, The Netherlands\\
\email{tksmit@protonmail.com}\\
\email{\{x.lu,h.a.reijers\}@uu.nl}
}
\maketitle              
\begin{abstract}
    Predictive Process Monitoring focuses on predicting future states of ongoing process executions, such as forecasting the remaining time. Recent developments in Object-Centric Process Mining have enriched event data with objects and their explicit relations between events. 
    To leverage this enriched data, we propose the Heterogeneous Object Event Graph encoding (HOEG), which integrates events and objects into a graph structure with diverse node types. It does so \emph{without aggregating} object features, thus creating a more nuanced and informative representation.  
    We then adopt a heterogeneous Graph Neural Network architecture, which incorporates these diverse object features in prediction tasks.
     We evaluate the performance and scalability of HOEG in predicting remaining time, benchmarking it against two established graph-based encodings and two baseline models.
    Our evaluation uses three Object-Centric Event Logs (OCELs), including one from a real-life process at a major Dutch financial institution. The results indicate that HOEG competes well with existing models and surpasses them 
    when OCELs contain \tocheck{informative object attributes and event-object interactions}.
    
\end{abstract}
\section{Introduction} 
\label{sec:intro}

Predictive Process Monitoring (PPM) focuses on predicting the future status of ongoing process executions through historical event data. PPM techniques help to monitor ongoing process executions and offer timely interventions. 
Traditionally, PPM has relied on single-case event logs as the primary data source~\cite{teinema2019-outcome_ppm_survey}. However, recent advancements in Object-Centric Process Mining (OCPM) have begun to enhance these event logs by incorporating objects and their interactions with events, providing a richer data context for training predictive models. 

Early attempts in object-centric PPM have shown promising results~\cite{adams2022-framework,galanti2023-ocppm,gherissi2023-ocppm,rohrer2022-thesis}.
These studies have introduced novel methods of feature extraction and encoding from an object-centric perspective. Adams et al.~\cite{adams2022-framework} proposed event-level encoding using object-centric directly-follows graphs (OC-DFGs)~\cite{adams2022-framework}, while Berti et al.~\cite{berti2022-graph_features_OCEL} focused on object-level encoding, leveraging object interactions via object graphs. Despite these advancements in handling multi-object data in OCELs, challenges remain in how to effectively encode object features and integrate them for prediction tasks.

To illustrate, consider a scenario involving process executions with object attributes \texttt{route length} and \texttt{urgency}. The challenge arises when these varying object attributes need to be integrated into an event-level encoding. For instance, including \texttt{route length} as an event feature can be problematic if it only applies to some events, potentially confusing the model due to missing values~\cite{galanti2023-ocppm}.
Similarly, encoding an object attribute like \texttt{urgency} as an event feature requires aggregating (e.g., averaging) the values across multiple objects, leading to information loss and less precise predictions, especially when there is significant variability in urgency levels~\cite{adams2022-framework,galanti2023-ocppm}.
%
%
Similar to event-level encoding, the object graph approach~\cite{berti2022-graph_features_OCEL} requires aggregating events when encoding them at the object level. Such an aggregation leads to a loss of critical information.

This indicates a need for a more integrated approach that considers both event and object perspectives \emph{natively}, without resorting to data manipulation techniques such as flattening or aggregating.
To address this challenge, our research focuses on developing a general approach for encoding object-centric events, objects, and their attributes natively, aimed at enhancing prediction performance. We propose the Heterogeneous Object Event Graph encoding (HOEG\footnote{Open source implementation is found at \url{https://github.com/TKForgeron/OCPPM}.}) that integrates events and objects into a graph with diverse node types without \emph{aggregating} object attributes. More specifically, HOEG encodes different types of nodes to represent events and objects, creating a more intricate graph structure. 

%
%
To assess the efficacy of HOEG, we measured the performance and runtime of HOEG. We compared these with two existing graph-based encodings and two baseline models in predicting the remaining time of a process. To allow for this assessment, we implemented a heterogeneous Graph Neural Network (GNN) architecture, adapted from a homogeneous counterpart, to optimize object information integration in our predictions.
We conducted experiments using three object-centric event logs, including one from a real-life process in a major Dutch financial institution. Our results demonstrate that HOEG outperforms existing approaches when the OCEL contains informative object attributes and interactions. It performs similarly to the other approaches when object features or interactions are less informative.


Our contribution to the field of System Engineering lies in providing an advanced method for encoding event and object data in predictive tasks. The implementation of HOEG is specifically tailored for enhanced and native encoding and analysis, aiming to yield more accurate predictive models for OCELs. Such developments are essential for training more sophisticated and reliable models in complex information systems.

\ch{
The remainder of this paper is organized as follows. First, we discuss related work and preliminaries in Section~\ref{sec:relatedwork} and Section~\ref{sec:preliminaries}, respectively. Next, we introduce our approach in Section~\ref{sec:approach}, the evaluation in Section~\ref{sec:eval}, and the results in Section~\ref{sec:results}. The discussion is presented in Section~\ref{sec:discussion}. Finally, we conclude the paper in Section~\ref{sec:conclusion}.
}

\section{Related Work} 
\label{sec:relatedwork}
We discuss five related works that used OCELs for PPM tasks. \autoref{tab:characterization_ocppm_works} lists the related works, compared along six aspects of their encodings. 

\begin{table}[b!]
\centering
\caption{Characterization of encoding approaches in related PPM works on OCELs.}
\label{tab:characterization_ocppm_works}
\resizebox{\textwidth}{!}{%
\begin{tabular}{lllllll}
\hline
\textbf{Approach} & \textbf{\begin{tabular}[c]{@{}l@{}}Encoding \\ Granularity\end{tabular}} &\textbf{\begin{tabular}[c]{@{}l@{}}Process \\ Behavior \\ Encoding\end{tabular}} & \textbf{\begin{tabular}[c]{@{}l@{}}Event \\ Attribute \\ Encoding\end{tabular}} & \textbf{\begin{tabular}[c]{@{}l@{}}Event-object \\ Interaction \\ Encoding\end{tabular}} & \textbf{\begin{tabular}[c]{@{}l@{}}Object-object \\ Interaction \\ Encoding\end{tabular}} & \textbf{\begin{tabular}[c]{@{}l@{}}Object \\ Attribute \\ Encoding\end{tabular}} \\ \hline

\textbf{Non-native} &  &  &  &  & \\
\textit{Rohrer et al.}~\cite{rohrer2022-thesis} & Event-level & One-hot & $-$ & $-$ & $-$ & For one object type only \\ 
\textit{Gherissi et al.}~\cite{gherissi2023-ocppm} & Event-level & One-hot & $-$ & Event object type count & $-$ & $-$ \\
\hdashline
\textbf{Native} &  &  &  &  &  & \\
\textit{Galanti et al.}~\cite{galanti2023-ocppm} & Event-level & CatBoost & Categorical & Event object type count & $-$ & Aggregated \& filled object attribute \\
 &  & algorithm & event attributes & Aggregated activity-object relation & & Aggregated previous object attribute \\
\textit{Berti et al.}~\cite{berti2022-graph_features_OCEL} & Object-level & $-$ & $-$ & Activity count & Object graph & $-$ \\
 &  &  &  & Related events count & features & \\
\textit{Adams et al.}~\cite{adams2022-framework} & Event-level & OC-DFG & Event-based & Current total object count & $-$ & $-$ \\
 &  & data structure & features & Previous object count &  &  \\
 &  &  & & Previous object type count & & \\
 &  &  & & Event objects & & \\
 &  &  & & Event object count & & \\
 &  &  & & Event object type count & & \\
 \hdashline
\textbf{HOEG} &  &  &  &  &  & \\
\textit{This work} & Event- and & OC-DFG & Event-based & Directly encoded & Direct encoding & Directly encoded \\
 & object-level & data structure & features & in the heterogeneous graph & into graph &  \\
 &  &  &  & data structure & structure possible & \\
\hline
\end{tabular}
}
\end{table}

Both Gherissi et al.~\cite{gherissi2023-ocppm} and Rohrer et al.~\cite{rohrer2022-thesis} followed a \emph{non-native} approach, meaning they followed the idea of flattening an OCEL into a classical event log, \tocheck{by performing lossy data preprocessing techniques.} Due to the flattening approach, the attributes of other objects are also not encoded. 
Interestingly, they also argued that OCELs allow for better predictions in PPM when compared to traditional event logs.

Three other approaches \cite{adams2022-framework,berti2022-graph_features_OCEL,galanti2023-ocppm} propose to use OCEL-\emph{native} feature extraction techniques. This means OCELs are \emph{not flattened} along one object type during preprocessing but an object-centric case notion is used. 
Berti et al.~\cite{berti2022-graph_features_OCEL} extract features per object, generating object vectors. As a consequence, their approach cannot be used for event-level prediction. In addition, due to this object-level view on OCELs, it cannot encode event-level features without losing information.
On the contrary, Adams et al.~\cite{adams2022-framework} design their features on an event basis, enriching event vectors. However, they are unable to encode diverse object attributes and interactions without resorting to aggregations.
Therefore, both works take a more methodical approach, describing a set of \emph{aggregated} features for \emph{event-object interactions} that can be applied to any OCEL, listed in~\autoref{tab:characterization_ocppm_works}(Column~5). 

Galanti et al.~\cite{galanti2023-ocppm} followed the \emph{native} approach and encoded OCELs at the event level while enabling the inclusion of object features. However, this method requires a predefined event vector structure, which consequently requires the \emph{aggregation} of values or the \emph{imputation} of missing values across all events. As a result, they do not support GNNs but only models that autofill missing values, like CatBoost.



Synthesizing, we observe information loss through \emph{flattening} of OCELs, \emph{aggregation} of object features, and \emph{imputation} of missing values. These limitations are caused by the inherently complex structure of OCELs. 
%
When creating a feature vector per event~(event-level granularity), this vector can refer to multiple objects with diverse attributes. 
To obtain an equal-size vector for each event, either none, a (manually) filled, or an aggregate of the object attributes (or other object-based features) may be taken into the vector. 
%
Similarly, when taking an object perspective, one creates a feature vector per object, trying to encode all events, leading to information loss~\cite{berti2022-graph_features_OCEL}).
In both perspectives, we experience a loss of information due to the unavailability of attributes or taking aggregates.

In this paper, we propose an encoding approach for GNN that captures object-centric event data without flattening the log, aggregating object features, or filling unavailable object features.
The proposed encoding provides support for all six dimensions presented in Table \ref{tab:characterization_ocppm_works}. \ch{For a detailed explanation and discussion of each aspect listed in Table~\ref{tab:characterization_ocppm_works}, we direct the reader to~\cite[p.~32]{smit2023}.}

%
%

\section{\ch{Preliminaries}} 
\label{sec:preliminaries}

In this paper, we adapt the definitions of event log, object graph, process execution, and execution extraction that are introduced in~\cite{adams2022-cases_variants_oced}. For the sake of clarity and completeness, we provide a concise overview of these concepts in this section. Note that the definitions of event logs with objects are generic and also support the formalizations defined in~\cite{DBLP:books/sp/22/Fahland22}. 

Given a set $X$, we use $\sigma = \langle x_1, x_2, \cdots, x_n \rangle$ to refer to a a sequence of length $n$ over $X$. 
For an element $x \in X$ and a sequence $ \sigma \in X^*$, we overload the notation $x \in \sigma$, expressing the occurrence of element $x$ in the sequence $x \in \text{range}(\sigma)$.

To define event logs, we introduce the following universe: $\mathcal{E}$ denotes the universe of event identifiers, $\mathcal{D}$ denotes the universe of attributes and $\mathcal{V}$ denotes the universe of attribute values.
In this paper, we are dealing with event data of different object types. 
Therefore, $\mathcal{T}$ defines the universe of types. There can be multiple instantiations of one type. We refer to each instantiation as an object. 
$\mathcal{O}$ defines the universe of objects. Each object is of one type $\text{type} : \mathcal{O} \rightarrow \mathcal{T}$. 
We define an event log that contains events and objects of different types, assigning each object to an event sequence. 

\begin{definition}[Event Log with Objects] An event log $L=(E,O,\sigma, \pi_e, \pi_o)$ is a tuple where:
    $E \subseteq \mathcal{E}$ is a set of events;
    $O \subseteq \mathcal{O}$ is a set of objects;
    $\sigma : \mathcal{O} \rightarrow E^*$ maps each object to an event sequence by overloading $\sigma$;
    $\pi_e : E \times \mathcal{D} \rightarrow \mathcal{V}$ maps event attributes onto values;
    $\pi_o : O \times \mathcal{D} \rightarrow \mathcal{V}$ maps object attributes onto values;
\end{definition}

\noindent For example, Table~\ref{tab:ocel_events_otc_example} shows an example of such an event log where the set of events $E = \{e_1, \cdots, e_{10}\}$ and the set of objects $O = \{o1, o2, i1, i2, i3, p1, p2, d1\}$. For instance, we have $\sigma(d1) = \langle e_9, e_{10}\rangle$, and $\pi_e(e1, \texttt{time}) =$``2023-01-30''. Table~\ref{tab:ocel_objects_otc_example} lists the object attributes, e.g., $\pi_o(o1, \texttt{Urgency}) = 1.0$.

Furthermore, we define the following two notations for $L=(E,O,\sigma, \pi_e, \pi_o)$: 
%
    (1) $\mathit{obj}(e)$ denote the object(s) associated to event $e \in E$, i.e., $\mathit{obj}(e) = \{o \in O | e \in \sigma(o)\}$; 
    (2) $\mathit{conn}_L$ defines the directly-follows relationships for all events and all objects, i.e., $\mathit{conn}_L = \cup_{o\in O} \cup_{1\leq i < \mid \sigma(o) \mid} \{(e_i,e_{i+1}) \in E \times E | e_i,e_{i+1} \in \sigma(o) \}$. 

We define the relationships between objects by defining the object graph of a log. 

\begin{definition}[Object Graph] Let  $L=(E,O,\sigma, \pi_e, \pi_o)$  be an event log. The object graph is an undirected graph $ OG_L = (O, C_O) $ with the set of undirected edges $C_O =\{ \{o_1, o_2\} \subseteq O | \exists e \in E :  o_1, o_2 \in \text{obj}(e) \wedge o_1 \neq o_2 \}$. 
\end{definition}


\begin{definition}[Process Execution]\label{def:pexecution}
Let $L=(E,O,\sigma, \pi_e, \pi_o)$ be an event log and \( O' \subseteq O \) be a subset of objects that forms a connected subgraph in \( OG_L \). The process execution of \( O' \) is a directed graph \( p_{O'}=(E', D) \) where
\begin{itemize}
    \item 
    \( E' = \{ e \in E \,|\, O' \cap \text{obj}(e) \neq \varnothing \} \) are the nodes, and
    \item 
    \( D = \text{conn}_L \cap (E' \times E') \) are the edges.
\end{itemize}
\end{definition}

Given the object graph, we may extract each connected component $(O', C'_O)$ of the graph and obtain the process execution $p_{O'} = (E', D')$.

\begin{definition}[Connected Component Extraction] \label{def:ccextraction}
Let  $L=(E,O,\sigma, \pi_e, \pi_o)$  be an event log and $OG_L = (O, C_O)$ its object graph. 
Function $\mathit{ext}_{\mathit{comp}}(L) = \{ p_{O'} \,|\, \allowbreak (O' \subseteq O) \wedge (O', C_O \cap (O' \times O')) \text{ is a connected component of } OG_L\}$ extracts process executions by connected components.
\end{definition}
Note that there are other extraction functions, for example, the leading type extraction defined in~\cite{adams2022-cases_variants_oced}.

\section{HOEG Approach} 
\label{sec:approach}

\begin{table}[tb!]
\centering
\caption{Events table of example log.}
\label{tab:ocel_events_otc_example}
\resizebox{\textwidth}{!}{%
\begin{tabular}{r c ccccccc}
\hline
\textbf{ID} & & \textbf{Activity} & \textbf{Time} & \textbf{Resource} & \textbf{Order} & \textbf{Item} & \textbf{Package} & \textbf{Delivery} \\ \hline
e1 & & Place order & 2023-01-30 & CloudServiceA & \{o1\} & \{i1, i2\} & \{\} & \{\} \\
e2 & & Pay order & 2023-01-30 & CloudServiceA & \{o1\} & \{\} & \{\} & \{\} \\
e3 & & Place order & 2023-01-30 & CloudServiceB & \{o2\} & \{i3\} & \{\} & \{\} \\
e4 & & Pay order & 2023-01-30 & CloudServiceB & \{o2\} & \{\} & \{\} & \{\} \\
e5 & & Pick item & 2023-01-31 & WarehouseTeamX & \{o1\} & \{i1\} & \{\} & \{\} \\
e6 & & Pick item & 2023-01-31 & WarehouseTeamX & \{o2\} & \{i3\} & \{\} & \{\} \\
e7 & & Pack item & 2023-01-31 & WarehouseTeamX & \{o1\} & \{i1\} & \{p1\} & \{\} \\
e8 & & Pack item & 2023-01-31 & WarehouseTeamX & \{o2\} & \{i3\} & \{p2\} & \{\} \\
e9 & & Ship package & 2023-02-01 & WarehouseTeamY & \{o1, o2\} & \{i1, i3\} & \{p1, p2\} & \{d1\} \\
e10 & & Confirm delivery & 2023-02-02 & PostalServiceP & \{o1, o2\} & \{i1, i3\} & \{p1, p2\} & \{d1\} \\
\hline
\end{tabular}%
}
%
\vspace{1em}
\centering
\caption{Object tables per different object type in the example log.}
\label{tab:ocel_objects_otc_example}
\resizebox{\textwidth}{!}{%
\begin{tabular}{ccc l ccc l cccc l cccc}
\cline{1-3} \cline{5-7} \cline{9-12} \cline{14-17}
\textbf{ID} & & \textbf{Urgency} & \multicolumn{1}{c}{$\ \ $} & \textbf{ID} & & \textbf{Discount} & \multicolumn{1}{c}{$\ \ $} & \textbf{ID} & & \textbf{Weight} & \textbf{Size} & \multicolumn{1}{c}{$\ \ $} & \textbf{ID} & & \textbf{Route length} & \textbf{No. stops} \\ \cline{1-3} \cline{5-7} \cline{9-12} \cline{14-17}
o1 & & 1.0 &  & i1 & & 33.0 &  & p1 & & 3.5 & medium &  & d1 & & short & 5.0 \\ \cline{14-17} 
o2 & & 3.0 &  & i2 & & 0.0 &  & p2 & & 3.0 & medium &  &  &  &  &  \\ \cline{1-3} \cline{9-12}
 & &  &  & i3 & & 25.0 &  &  &  &  &  &  & \multicolumn{1}{c}{} &  &  &  \\ \cline{5-7}
\end{tabular}%
}
\end{table}


As discussed, to be able to encode OCELs \emph{natively} both at event- and object-level, to leverage diverse object types and their features without aggregation or the need \tocheck{for imputing missing values}, we present the HOEG approach in the following.

The HOEG approach incorporates both event-level and object-level perspectives into a graph using different node and edge types. In a heterogeneous graph, each node type may have its shape, equal to how OCELs contain data entities that can each have a different shape~\cite{ghahfarokhi2020-ocel}. This enables HOEG to directly encode the complex relationships found in OCELs so that each relationship or attribute is left fully intact. 

\begin{definition}[HOEG]
\label{def:hoeg}
Let $L=(E,O,\sigma, \pi_e, \pi_o)$ be an object-aware log and $P = \{p_{O_1}, \allowbreak \cdots, p_{O_k}\}$ the set of extracted process executions. 
For each execution $p_{O} = (E_{O}, D_{O}) \in~P$, 
HOEG composes the related events and objects into a heterogeneous execution graph.
%
We define each heterogeneous execution graph $HOEG \in \mathcal{HOEG}$ as the tuple $HOEG=\left( \mathcal{NT}, \mathcal{ET}, \mathcal{X}, \mathcal{A}, f_{ntflookup}, f_{etflookup}, f_{etalookup} \right)$, with:
\begin{enumerate}[itemsep=0pt]
    \item[-] $\mathcal{NT}$ is the set of node types in HOEG. Each node type $\mathit{NT} \in \mathcal{NT}$ represents a semantically distinct group of nodes. $\mathcal{NT}$ must include node type \text{\texttt{event}} \ch{and at least one object type}, i.e., $\abs{\mathcal{NT}} \geq 2$.
    \item[-] $\mathcal{ET}$ is the set of edge types in HOEG. Each edge type $\mathit{ET} \in \mathcal{ET}$ describes a type of relationship between nodes via a semantic triple $( \ssubject, \spredicate, \allowbreak \sobject)$, such that $\ssubject, \sobject \in \mathcal{NT}$ and $\spredicate$ equals any word that describes the type of interaction the nodes have. 
    \item[-] $\mathcal{X}$ defines a set of feature matrices $\{X_1,\dots,X_m\}$. For each node type $NT_i \in \mathcal{NT}$, there exists a feature matrix $X_i$, such that $X_i$ is a matrix representing features associated with nodes of type $NT_i$.
    Optionally, for each edge type $ET_j \in \mathcal{ET}$, there may also be feature matrix $X_j \in \mathcal{X}$, such that $X_j$ represents a matrix of edge features associated with edges of type $ET_j$, \ch{i.e., $\abs{\mathcal{X}} \geq \abs{\mathcal{NT}}$.} 
    \item[-] $\mathcal{A}$ defines a set of adjacency matrices $\{A_1,\dots,A_v\}$.
    For each edge type $ET_i \in \mathcal{ET}$, there exists an adjacency matrix $A_i \in \mathcal{A}$, such that $A_i$ is a matrix representing connections between nodes based on edge type $ET_i$.
    \item[-] Lookup functions $f_{ntflookup}$, $f_{etflookup}$, and $f_{etalookup}$ map node and edge types to their corresponding feature and adjacency matrices:
        (1)~$f_{ntflookup}: \mathcal{NT} \rightarrow \mathcal{X}$;
        \ch{(2)~$f_{etflookup}: \mathcal{ET}  \pfun \mathcal{X} $};
        (3)~$f_{etalookup}: \mathcal{ET} \rightarrow \mathcal{A}$.
\end{enumerate}
An OCEL that is encoded via HOEG can be expressed as the set $\mathcal{HOEG}=\{ HOEG_1, \dots,\allowbreak HOEG_k \}$, respectively. 
\end{definition}

\begin{figure}
    \centering
    \includegraphics[width=\textwidth]{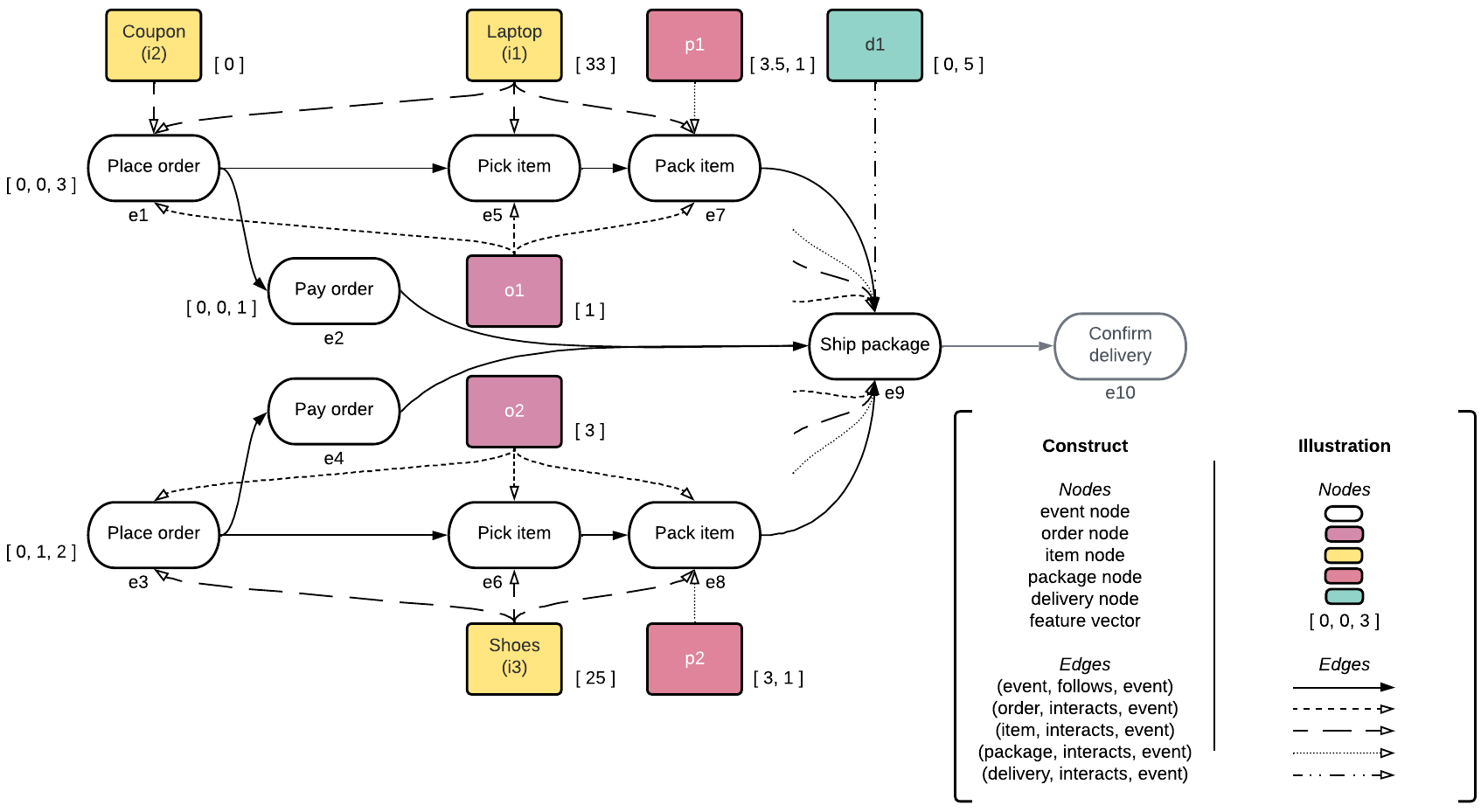}
    \caption{Heterogeneous object event graph for Execution A (derived from Tables \ref{tab:ocel_events_otc_example} and \ref{tab:ocel_objects_otc_example}) of the running OTC example. \textit{Note}: the trimmed (for readability) edges going out from \textit{Ship package} (\texttt{e9}) are connected with objects \texttt{o1}, \texttt{o2}, \texttt{i1}, \texttt{i3}, \texttt{p1}, \texttt{p2}. Faded event node \texttt{e10} signifies future event \textit{Confirm delivery}. Also note that for readability, not all edges are drawn. 
    \label{fig:HOEG_concept}
    }
\end{figure}

As an example, consider a process execution (Execution A) of an Order-to-Cash process resulting in an example log listed in Table~\ref{tab:ocel_events_otc_example}, and the referenced objects listed in Table~\ref{tab:ocel_objects_otc_example}. Figure~\ref{fig:HOEG_concept}, then, visualizes one HOEG instance.

HOEG in Figure~\ref{fig:HOEG_concept} uses connected components extraction (see Def.~\ref{def:ccextraction}) to construct the OC-DFG that relates the events. Objects are connected to events via the references found in the OCEL (cf. Table~\ref{tab:ocel_events_otc_example}).
Let us express Figure~\ref{fig:HOEG_concept} in terms of Definition~\ref{def:hoeg}. 
\\
$HOEG_{Execution_A}=\left( \mathcal{NT}, \mathcal{ET}, \mathcal{X}, \mathcal{A}, f_{ntflookup}, f_{etflookup}, f_{etalookup} \right)$, where:
\begin{align*}
\mathcal{NT} = \{\text{\texttt{event}, \texttt{order}, \texttt{item}, \texttt{package}, \texttt{delivery}}\}
    \end{align*}%
    \vspace{-2em}
\begin{align*}
\mathcal{ET} = \{   & (\text{\texttt{event}, \texttt{follows}, \texttt{event}}),
                    (\text{\texttt{order}, \texttt{interacts}, \texttt{event}}), \\
                    & (\text{\texttt{item}, \texttt{interacts}, \texttt{event}}),
                    (\text{\texttt{package}, \texttt{interacts}, \texttt{event}}), \\
                    & (\text{\texttt{delivery}, \texttt{interacts}, \texttt{event}}) \}
\end{align*}
For brevity and clarity, we represent only matrix dimensions of the elements in $\mathcal{X}$ and $\mathcal{A}$ paired with respective node and edge types. \\
The feature matrices $\mathcal{X}$ consist of:
\begin{enumerate}
    \item[-] \texttt{event}: $3 \times 9$, encoding events \{\texttt{e1}, \dots, \texttt{e9}\} with $3$ features.
    \item[-] \texttt{order}: $1 \times 2$, encoding objects \{\texttt{o1}, \texttt{o2}\} with attribute \texttt{urgency}.
    \item[-] \texttt{item}: $1 \times 3$, encoding objects \{\texttt{i1}, \texttt{i2}, \texttt{i3}\} with attribute \texttt{discount}.
    \item[-] \texttt{package}: $2 \times 2$, encoding objects \{\texttt{p1}, \texttt{p2}\} with attributes \texttt{weight} and \texttt{size}.
    \item[-] \texttt{delivery}: $2 \times 1$, encoding \{\texttt{d1}\} with \texttt{route length} and \texttt{no. stops}.
\end{enumerate}
The adjacency matrices $\mathcal{A}$:
\begin{enumerate}
    \item[-] $(\text{\texttt{event}, \texttt{follows}, \texttt{event}}): 2 \times 10$, encoding edges \{(\texttt{e1}, \texttt{e2}), (\texttt{e1}, \texttt{e5}), \dots, (\texttt{e8}, \texttt{e9})\}.
    \item[-] $(\text{\texttt{order}, \texttt{interacts}, \texttt{event}}): 2 \times 8$, encoding edges \{(\texttt{o1}, \texttt{e1}), \dots, (\texttt{o2}, \texttt{e9})\}.
    \item[-] $(\text{\texttt{item}, \texttt{interacts}, \texttt{event}}): 2 \times 9$, encoding edges \{(\texttt{i1}, \texttt{e1}), \dots, (\texttt{i3}, \texttt{e9})\}.
    \item[-] $(\text{\texttt{package}, \texttt{interacts}, \texttt{event}}): 2 \times 4$, encoding \{(\texttt{p1}, \texttt{e7}), \dots, (\texttt{p2}, \texttt{e8})\}.
    \item[-] $(\text{\texttt{delivery}, \texttt{interacts}, \texttt{event}}): 2 \times 1$, encoding \{(\texttt{d1}, \texttt{e9})\}.
\end{enumerate}

This example shows that, in comparison with the Event Feature Graph~(EFG) approach~\cite{adams2022-framework}, HOEG distinctly encodes objects and their features explicitly. Conversely, EFG tends to either lose object information or aggregate object information into event features (e.g., \textbf{O1-O6} in~\cite{adams2022-framework}).


\subsubsection{Machine Learning on HOEG}
When encoding OCELs using the HOEG approach, it enables the use of heterogeneous GNN architectures to perform machine learning tasks, as opposed to relying solely on homogeneous GNN architectures.
In heterogeneous GNN architectures, we can define different edge relation types over which the message-passing mechanism transfers information during a forward pass of a neural network. Via linear projection, it can handle the different feature matrix dimensions. Through this, we can train on both event and object features simultaneously. The network learns how to best combine and transform the information from the different node types (edge types) to predict a certain target. 
Therefore, the HOEG approach supports a flexible and expressive encoding for OCEL while embracing the inherent structure of OCELs.

\section{Evaluation} 
\label{sec:eval}
%
This section outlines the setup for evaluating the proposed HOEG. Firstly, we distinguish EFG as a graph-based encoding against which the HOEG is mainly compared. Their configuration is described. Secondly, GNN architecture design and hyperparameter tuning are discussed. Thirdly, evaluation methods are presented by which the graph-based feature encodings are compared and analyzed. Lastly, we explain the three OCELs used.

\subsection{Feature Encodings} \label{sec:feature_encodings}

We compare HOEG to the state-of-the-art graph-based encoding technique for OCELs: the EFG by Adams et al.~\cite{adams2022-framework}. \ch{Other approaches discussed in Section~\ref{sec:relatedwork} either do not support GNN natively or predict at the object level, making the results incomparable.} Table~\ref{tab:efg_hoeg_config} lists the configuration of both EFG and HOEG. 
It shows that HOEG complements EFG with diverse objects. \ch{Fundamentally, HOEG encodes the event nodes using the same features as EFG. Additionally, HOEG encodes each object type separately to include object attributes, also allowing more complex relations (event-object edges).} 
We set \textit{Remaining Time} (\textbf{P3} in \cite{adams2022-framework}) as the prediction target for both EFG and HOEG. 
\ch{This is calculated by subtracting the timestamp of the last event in a process execution from the current timestamp.
Using Z-score normalization, the target was standardized based on the training split, to mitigate information leakage.}

\begin{table}[tb]
\centering
\caption{Graph-based encoding technique configurations.}
\label{tab:efg_hoeg_config}
\resizebox{\textwidth}{!}{%
\begin{tabular}{lrr}
\hline
 & \multicolumn{1}{c}{\textbf{EFG}} & \multicolumn{1}{c}{\textbf{HOEG}} \\
 & \multicolumn{1}{c}{\textbf{(Adams et al. \cite{adams2022-framework})}} & \multicolumn{1}{c}{\textbf{(our approach)}} \\

\hline
\textbf{Process Behavior Encoding} & \multicolumn{1}{r:}{OC-DF graph structure} & OC-DF graph structure \\
 & \multicolumn{1}{r:}{} & enhanced with object nodes \\
 & \multicolumn{1}{l:}{} & \multicolumn{1}{l}{} \\
 
\textbf{Event Node Features} & \multicolumn{1}{r:}{C2, P2, P5, O3 in~\cite{adams2022-framework}} & C2, P2, P5, O3 in~\cite{adams2022-framework} \\
 & \multicolumn{1}{r:}{Numerically encoded event attributes} & Numerically encoded event attributes \\
 & \multicolumn{1}{l:}{} & \multicolumn{1}{l}{} \\

\textbf{Object Type Node Features} & \multicolumn{1}{r:}{N/A} & A node type per object type \\
 & \multicolumn{1}{r:}{} & with numerically encoded object attributes \\

\hline
\end{tabular}%
}
\end{table}


\subsection{Models and Hyperparameters}
To leverage the multi-dimensional information contained in OCELs, we use graph-based deep learning models for each of the given scenarios to demonstrate how the graph-based models exploit their respective structures. 

Though HOEG is agnostic of GNN layer choice, we employ k-dimensional GNN layers~\cite{morris2019-HigherOrderGNN} in our architecture design. The message-passing layer captures higher-order graph structures at multiple scales. Higher-order graph structures refer to patterns or relationships that exist between groups of nodes in a graph, beyond merely the pairwise connections between individual nodes. This mechanism helps capture the state of an event in its process execution graph.

In Table \ref{tab:hyperparams}, we outline the hyperparameters that are considered. Each row indicates the selected value or tunable value range. For the three experiments, we set nine hyperparameters to the recommended values via guidance from the literature. We choose the number of \textit{hidden dimensions} (\textit{hd}) and the \textit{learning rate} (\textit{lr}) to finetune in our first experiment to find the best value within a specified range, investigating the learning capability and stability of our approach.

\begin{table}[tb]
\centering
\caption{Hyperparameters used in the experiments.}
\label{tab:hyperparams}
\resizebox{\textwidth}{!}{%
\begin{tabular}{llll}
\hline
\textbf{Hyperparameter}         & \textbf{Value(s)}             & \textbf{Source(s)} & \textbf{Remark} \\ \hline
Batch size                      & $16$                          & & Dependent on available resources. \\
No. epochs                      & $30$                          & \cite{adams2022-framework} & \\
Early stopping criterion        & $4$                           & & Dependent on available resources. \\ 
No. pre-message-passing layers  & $0$                           & \cite{adams2022-framework} & We use preprocessed features. \\
No. message-passing layers      & $2$                           & \cite{adams2022-framework,morris2019-HigherOrderGNN} & k-Dimensional GNN\\
No. post-message-passing layers & $1$                           & \cite{adams2022-framework} & \\
Drop-out rate                   & $0.0$                         & \cite{you2020-design_space} & \\
Activation function             & PReLU                         & \cite{you2020-design_space} & \\
Optimizer                       & Adam                          & \cite{adams2022-framework}, \cite{you2020-design_space} & Using default settings. \\
No. hidden dimensions           &$\{8,16,24,32,48,64,128,256\}$ & & Adams et al.~\cite{adams2022-framework} used 24. \\
Learning rate                   & $\{0.01, 0.001\}$             & \cite{adams2022-framework}, \cite{you2020-design_space} & You et al.~\cite{you2020-design_space} recommend 0.01, \\
                                &                               & & which is also used in \cite{adams2022-framework}.\\
\hline
\end{tabular}%
}
\end{table}

\subsection{Baselines and Evaluation Metrics } \label{sec:eval_methods}
In addition to the EFG, three other baselines are used: 
%
(1) \textit{Median}, which is used as a lower-bound, 
(2)~\textit{LightGBM}, and (3)~\textit{EFG\textsubscript{ss}}. The \textit{LightGBM} is a lightweight gradient booster model, which seems to be promising for PPM tasks \cite{pourbafrani2022-rem_time_intercase}. Gradient boosting models require a tabular data structure with fixed-size dimensions. Therefore, the LightGBM baseline takes only a tabular version of the EFG as input data.

The \textit{EFG\textsubscript{ss}} is the baseline model where we replicate the EFG and GCN architecture~\cite{adams2022-framework} that uses subgraph sampling\footnote{The subgraph sampling is set to sample consecutive nodes of size four chronologically.}, which is a data augmentation technique. 
As this GNN baseline runs on a subgraphed EFG (configured equal to EFG in Table \ref{tab:efg_hoeg_config}), it uses a global pooling operation to make (sub)graph-level predictions.

Finally, to evaluate the performance, the mean absolute error (MAE) and mean squared error (MSE) metrics are calculated. To evaluate the scalability of our approach, the model training time and prediction time are measured in seconds. The experiment is run on Intel Core i5-7500 with 48GB RAM and NVIDIA GeForce GTX 960 (4GB).

\subsection{Object-Centric Datasets} \label{sec:intro_datasets}
We used three OCELs, listed in Table~\ref{tab:ocels_summary}. The first OCEL, BPI17, concerns a loan application process~\cite{bpic17}. \ch{Based on connected components extraction, a 56/14/30 split was taken as in \cite{adams2022-framework}.} The second OCEL, OTC, concerns an order management process. originally generated for demonstration purposes~\cite{aalst2020-discovering_OC_PNs,ghahfarokhi2020-ocel}, now also used for PPM experiments~\cite{gherissi2023-ocppm,rohrer2022-thesis}. \ch{Here, a 70/15/15 split was used, based on lead type (\textbf{item}) extraction because connected component extraction does not converge due to a high degree of object interaction (see Table \ref{tab:ocels_summary})}.

\begin{table}[tb!]
\centering
\caption{Summary of the datasets used in the experiments.}
\label{tab:ocels_summary}
\resizebox{0.8\textwidth}{!}{%
\begin{tblr}{
    colspec = {l cc ccc c c},
    column{1} = {gray9}, 
    column{4} = {gray9}, 
    column{5} = {gray9}, 
    column{6} = {gray9}, 
    column{8} = {gray9}, 
  }
\hline
\textbf{OCEL} & \textbf{Events} &  & \textbf{Objects} &  &  & \textbf{Cases} & \textbf{Mean Object} \\
 & & \textbf{Attributes} & & \textbf{Types} & \textbf{Attributes} & & \textbf{Interactions per Event} \\ 
\hline
\textbf{BPI17} & $393,931$ & $3$ & $74,504$ & $2$ & $10$ & $31,509$ & $1.35$ \\
\textbf{OTC} & $\ \: 22,367$ & $4$ & $11,484$ & $3$ & $\ \: 1$ & $\ \: 8,159$ & $7.15$ \\
\textbf{FI} & $695,694$ & $3$ & $94,148$ & $3$ & $14$ & $31,277$ & $1.09$ \\
\hline
\end{tblr}
}
\end{table}

The last OCEL, FI, is extracted from a real-life workflow management system (WfMS) operational at a large financial institution. The recorded process is part of the organization’s Know Your Customer liabilities. These include customer identification, risk assessment, due diligence, and monitoring. The data contains actual events executed in the WfMS and objects enriched with intelligence from other systems. The WfMS supports standardization but allows room for deviations. Altogether, this results in a comprehensive global process model, when object-centric process discovery is applied.
FI contains $94,148$ objects of three types that appear in sequential order throughout process executions: \textbf{signal}, \textbf{follow-up investigation}, and \textbf{complex follow-up investigation}. The three types are dossiers that progress in complexity, from a \textbf{signal} to a \textbf{complex follow-up investigation}. We extracted $695,694$ events recorded between January and May 2023. Events and objects were enriched with $3$ and $14$ attributes respectively. The OCEL includes complex process executions, that may start with events relating to a \textbf{signal} and end up with a series of \textbf{complex follow-up investigation} activities, while events are still being recorded relating to a \textbf{follow-up investigation} object. \ch{For model training, a 70/15/15 split was used, again based on connected components extraction.}

\section{Results} 
\label{sec:results}

This section discusses the results obtained through three experiments. In the first, we experiment with different hyperparameter settings, producing one best-performing model per encoding. The second experiment compares model performance across the encodings. Third, we elaborate on the baseline experiment, evaluating our models against different baselines. After this, interpretations are given that synthesize the experimental results into insights.

\subsection{Hyperparameter Tuning Experiment}
\begin{figure}[ht]
    \centering
    \includegraphics[width=\textwidth]{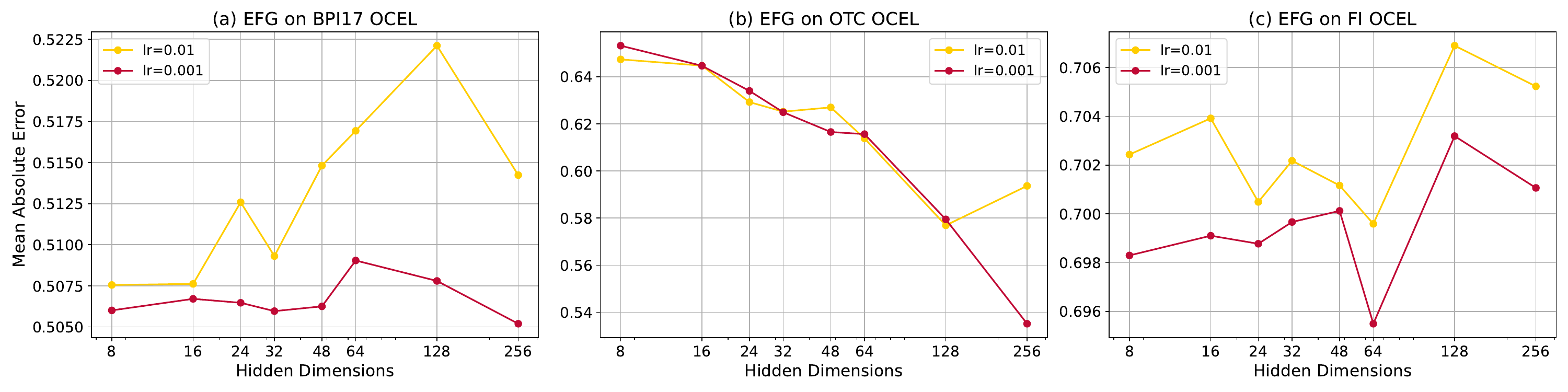}
    \includegraphics[width=\textwidth]{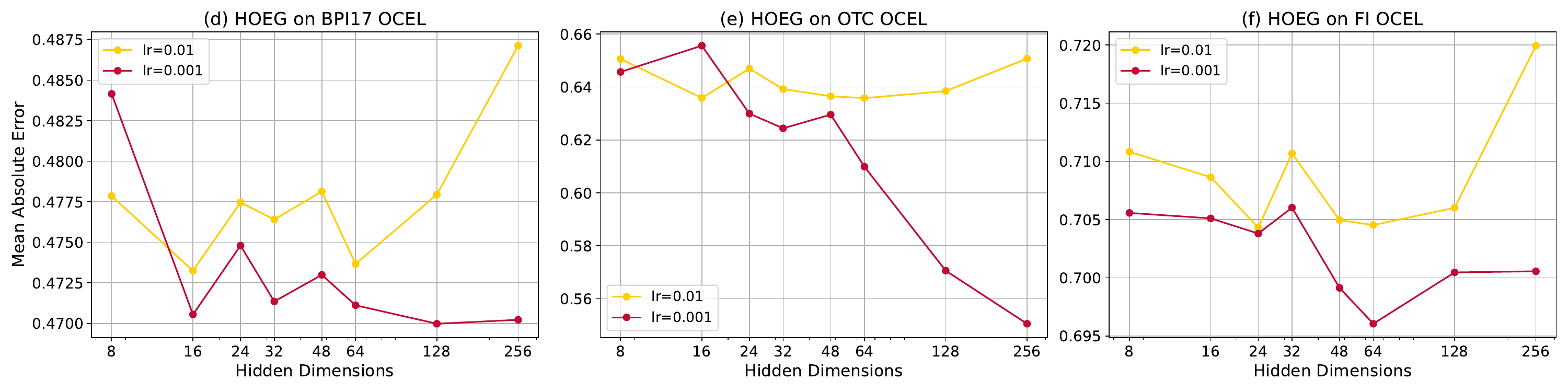}
    \caption{Test MAE scores for EFG and HOEG per hyperparameter setting. \textit{Note:} scales (y-axis) are not aligned, as we intend to compare hyperparameter settings within each encoding and dataset. }
    \label{fig:exp1_hp_tuning_efg_hoeg}
\end{figure}

Figure \ref{fig:exp1_hp_tuning_efg_hoeg} shows the effect of tuning \textit{learning rate} and \textit{hidden dimensions} simultaneously on the EFG-based and HOEG-based models.
A first observation shows that the models are stable for both BPI17 and FI, as the MAE score range is relatively small. Performance on OTC seems to be more susceptible to model configuration.

In terms of hyperparameters, we generally observe that lower \textit{learning rate} ($0.001$) scores better. Looking at \textit{hidden dimensions}, 256 seems to work best for BPI17 and OTC, while \textit{hidden dimensions} $=64$ yields the lowest MAE on the FI OCEL for both encodings. Next to that, we observe a pattern for the OTC dataset suggesting that increasing model complexity improves the learning capacity for both EFG and HOEG.

\subsection{Encoding Type Experiment}

\begin{figure}[htb]
    \centering
    \includegraphics[width=0.32\textwidth]{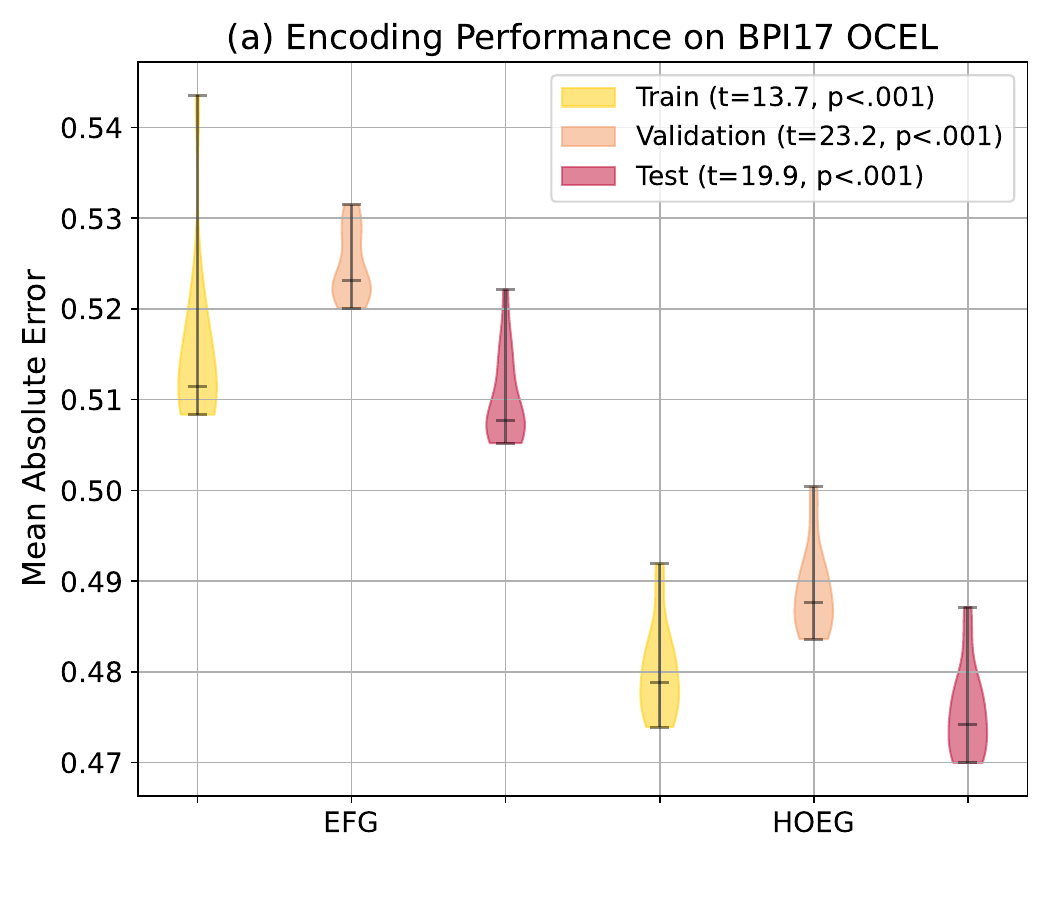}
    \includegraphics[width=0.56\textwidth]{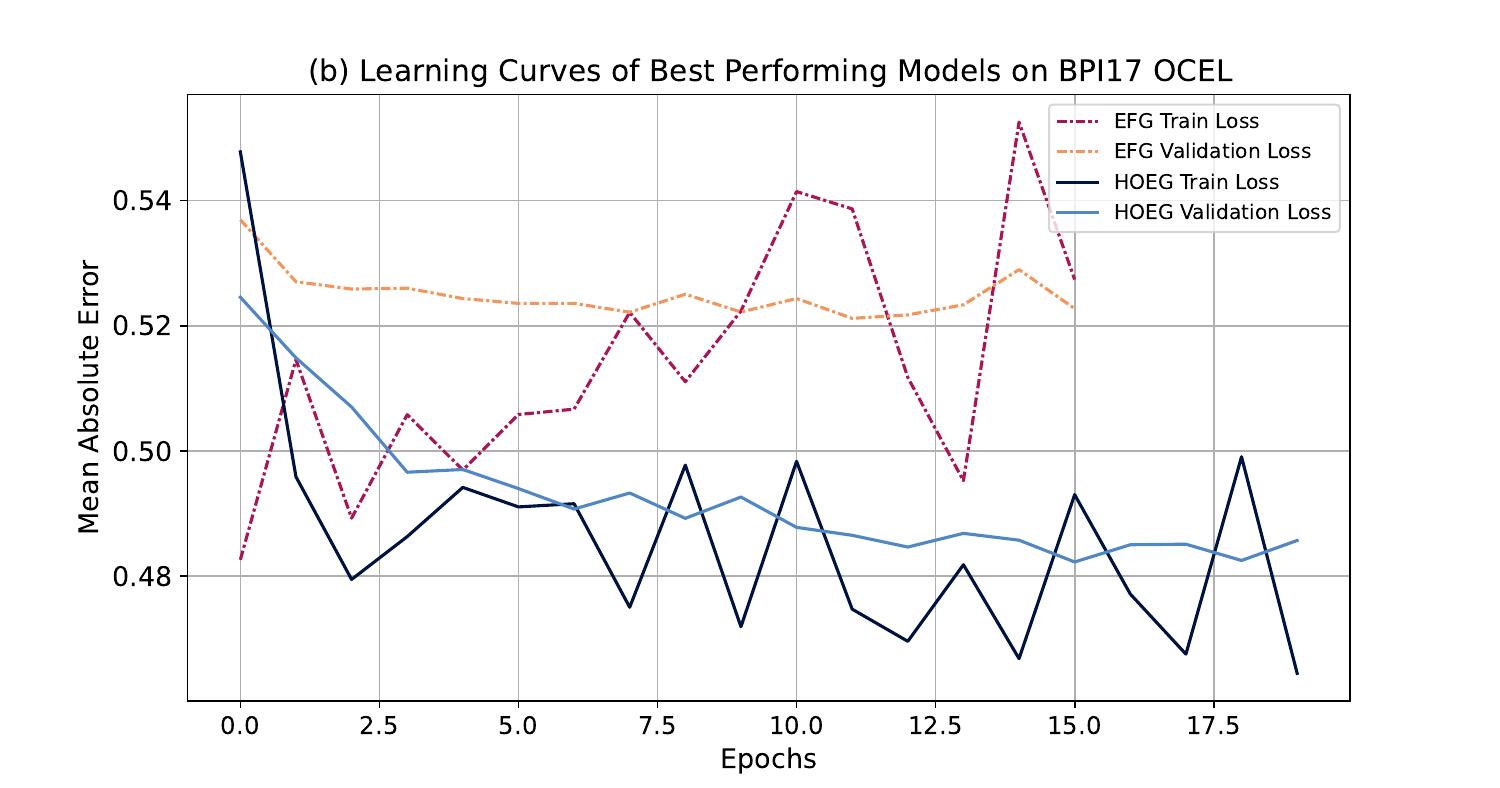}
    \includegraphics[width=0.32\textwidth]{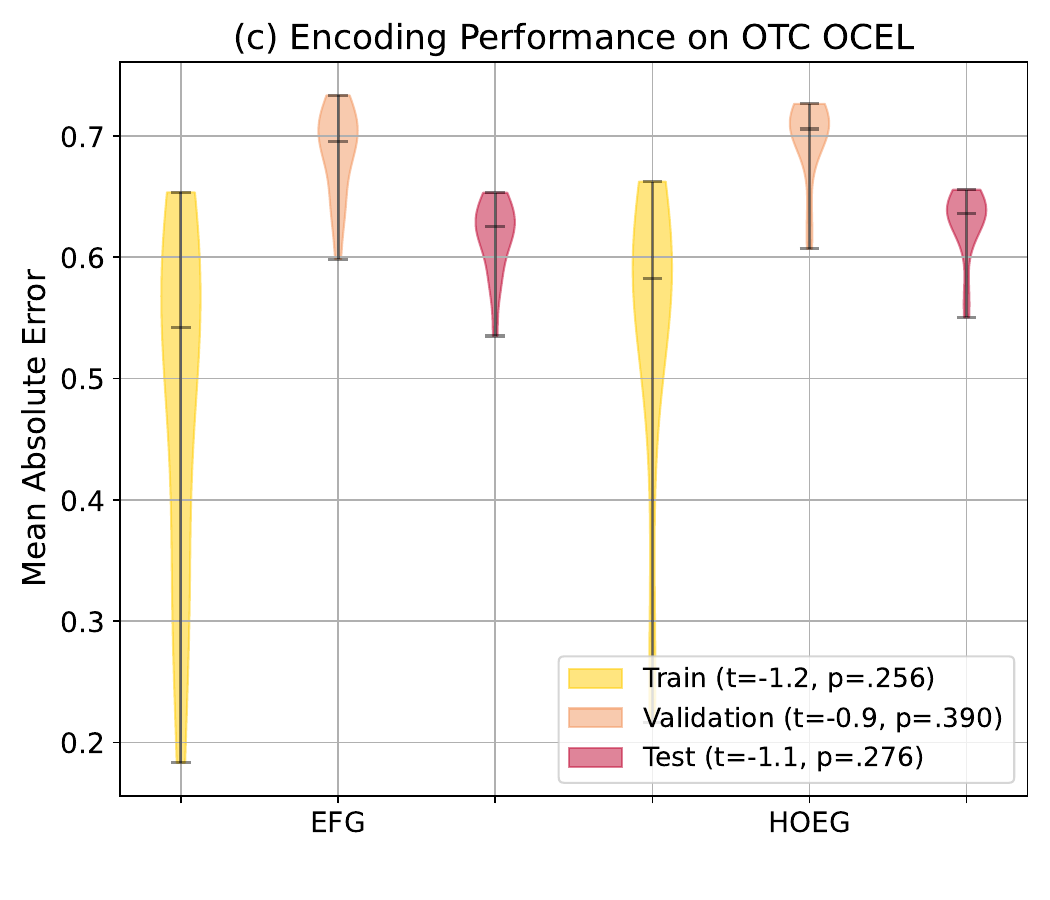}
    \includegraphics[width=0.56\textwidth]{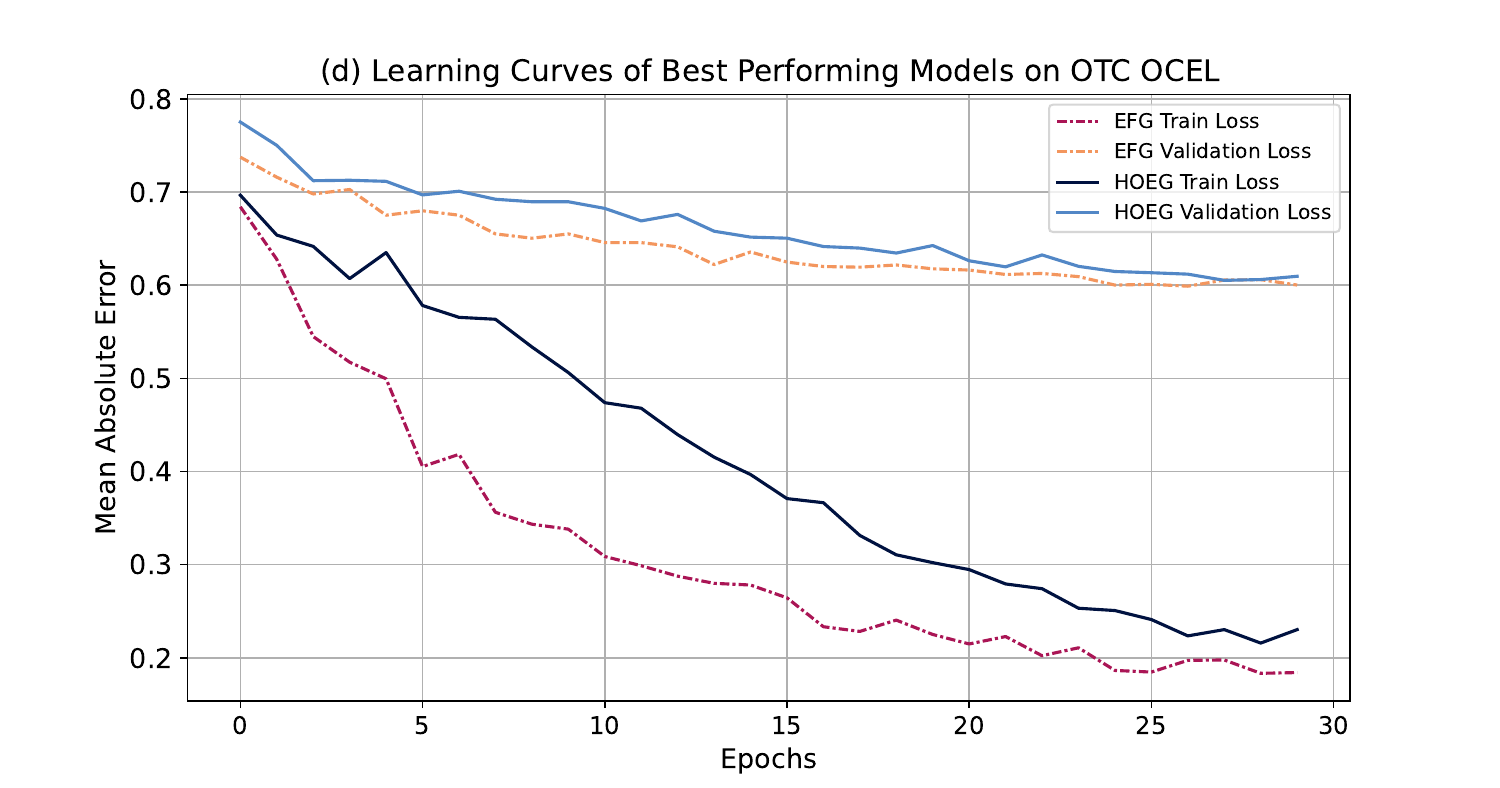}
    \includegraphics[width=0.32\textwidth]{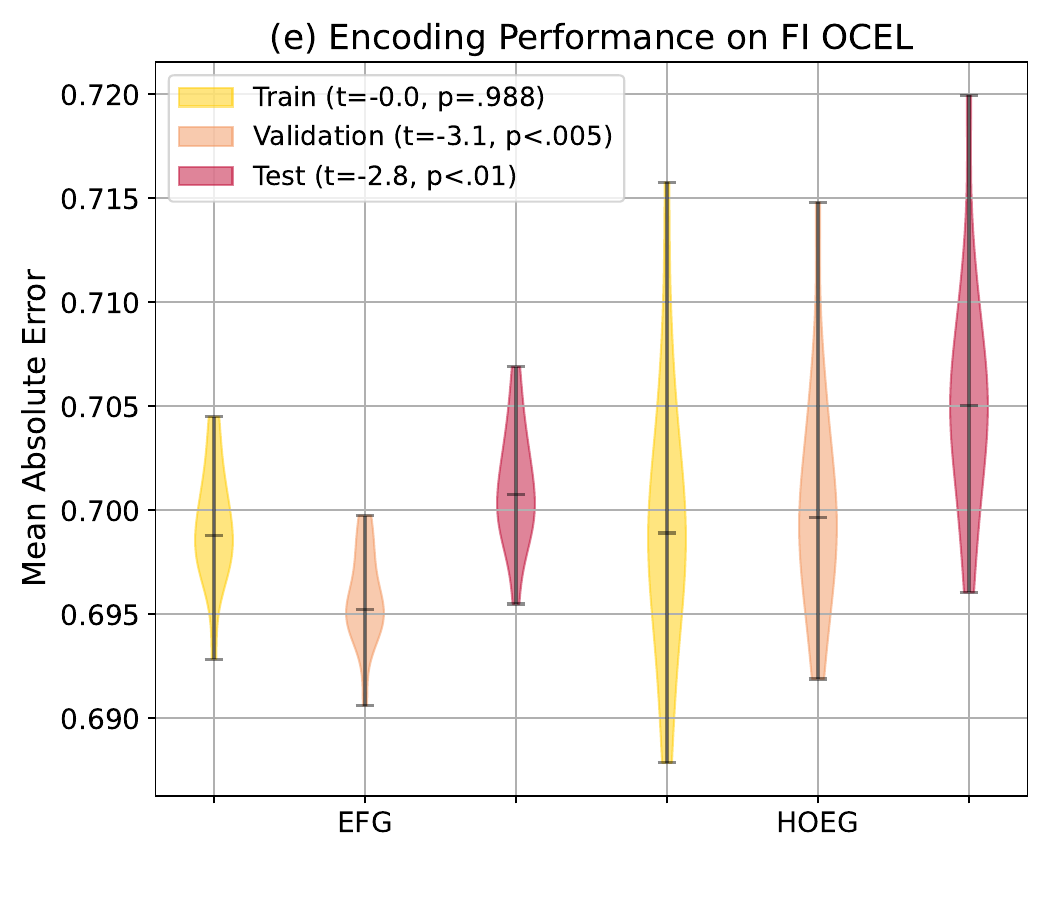}
    \includegraphics[width=0.56\textwidth]{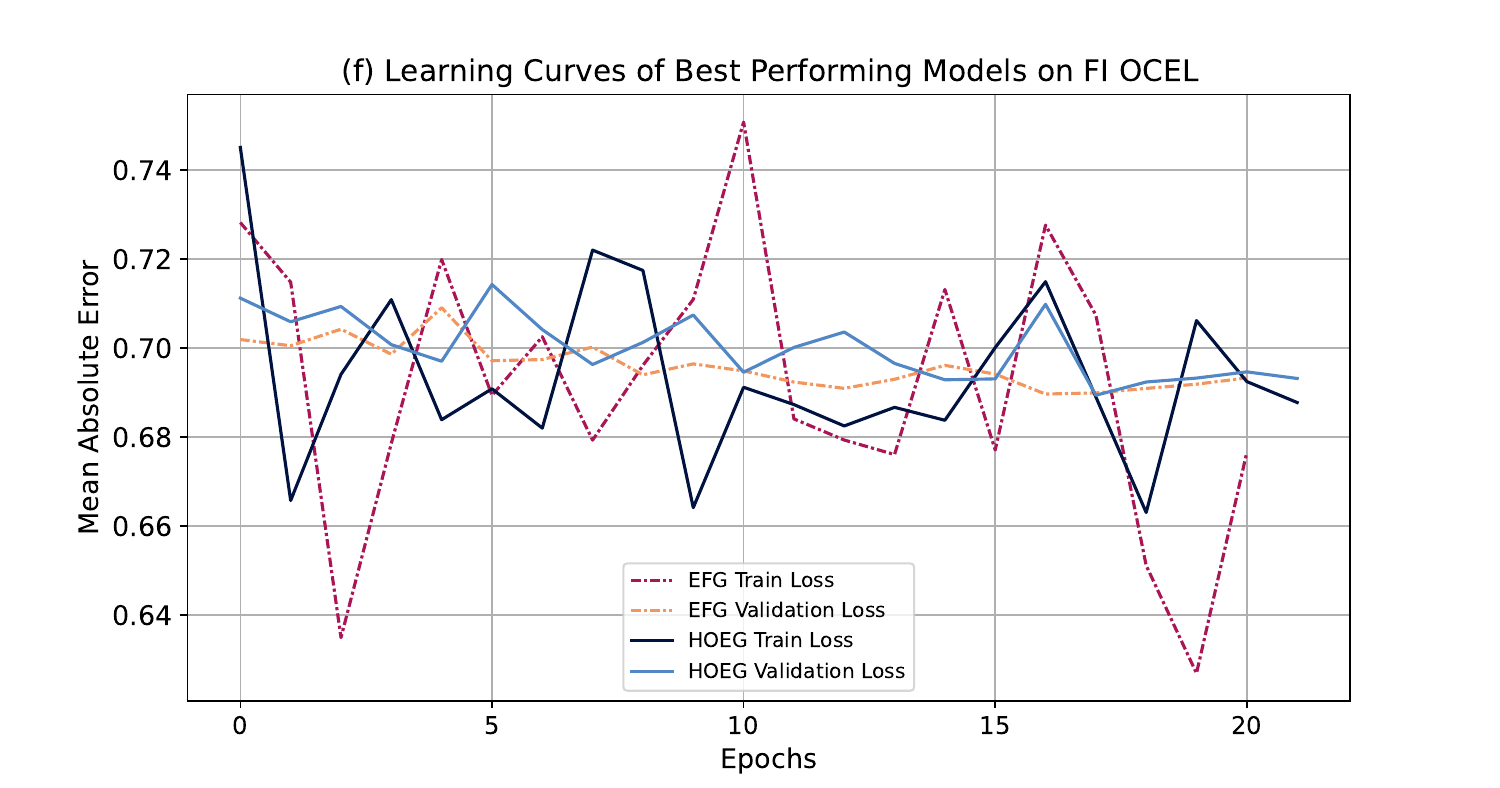}
    \caption{Violin plot of the MAE score distribution over the hyperparameter settings per split (a, c, e). Training and validation loss learning curves of tuned models (b, d, f). BPI17: (a) and (b), OTC: (c) and (d), FI: (e) and (f).}
    \label{fig:exp2}
\end{figure}

Figure \ref{fig:exp2} gives distributions of the model performance of all hyperparameter combinations per split (train, validation, test) for the three OCELs. \ch{To show statistical significance between EFG and HOEG performance (per split, $n=16$), a t-test was done.} In addition, the learning curves of the best models for both EFG and HOEG are plotted per dataset, providing insight into the training process and model performance during training. 

For the BPI17 OCEL, the results show that HOEG performs significantly better than EFG in terms of MAE and learning capacity. 
The learning curve of HOEG suggests relatively stable model performance. In contrast, the EFG curve exhibits some indications of instability, particularly on the training split. This is primarily attributed to a few outliers in the MAE scores on the training split. However, when considering the MAE scores across the three splits (violin plot), EFG's performance appears relatively consistent.

For the OTC OCEL, Figure \ref{fig:exp2} shows that both models achieve a similar MAE and show a stable learning process. However, both models score a worse MAE on the validation and test set, \ch{see Figure \ref{fig:exp2}(c)}. This suggests that the models seem to overfit on the training data and do not generalize well on the unseen data. This may be because the dataset does not contain realistic object-centric event data or contains too few examples.

For the FI OCEL, Figures \ref{fig:exp2}(e) and (f) also show that both models achieve a similar MAE. 
The learning curves signify that our GNNs struggle to learn relevant complex patterns for predicting the remaining time. Though HOEG’s learning curve presents itself as more stable, the validation loss is not going down by much throughout the epochs. This could be attributed to the additional data that HOEG includes compared to EFG (i.e., 14 object attributes). The violin plots show that EFG slightly outperforms HOEG on all hyperparameter configurations. Together with the learning curves, they suggest both models underfit this dataset.

The results of the second experiment are not conclusive about which encoding performs the best on OCEL logs for predicting the process remaining time. Looking at the learning curves, it might be suggested that HOEG trains in a more stable manner than EFG. 


\subsection{Baseline Experiment}

We compare HOEG to three other baseline models: the Median, LightGBM, and EFG\textsubscript{ss}~\cite{adams2022-framework}, in addition to EFG. By comparing the performance to the baseline models, the objective is to understand the relative impact of the results of the previous experiments. {Note that these are not tuned, hence the traces in the validation split are assigned to the train split.}

\begin{table}[tb]
\centering
\caption{Performance and scalability of HOEG and four baseline models. \textit{Note}: the best-performing hyperparameter configurations are used for EFG and HOEG. Also, EFG\textsubscript{ss} is italicized for reasons mentioned in Section~\ref{sec:discussion}. Blank validation scores imply there was no model tuning.}
\label{tab:baseline_table}
\resizebox{0.9\textwidth}{!}{%
\begin{tabular}{c l rr rr rr rr}
\hline
\textbf{OCEL} & \textbf{Model} & \multicolumn{2}{c}{\textbf{Train}} & \multicolumn{2}{c}{\textbf{Validation}} & \multicolumn{2}{c}{\textbf{Test}} & \multicolumn{1}{c}{\textbf{Fitting}} & \multicolumn{1}{c}{\textbf{Prediction}} \\
&  & \multicolumn{1}{c}{\textbf{MAE}} & \multicolumn{1}{c}{\textbf{MSE}} & \multicolumn{1}{c}{\textbf{MAE}} & \multicolumn{1}{c}{\textbf{MSE}} & \multicolumn{1}{c}{\textbf{MAE}} & \multicolumn{1}{c}{\textbf{MSE}} & \multicolumn{1}{c}{\textbf{Time (s)}} & \multicolumn{1}{c}{\textbf{Time (s)}} \\ 
\hline
\multirow{5}{*}{\rotatebox[origin=c]{90}{BPI17}} & \textbf{Median} & \color{darkgray}0.7854 & \color{darkgray}1.0802 &  &  & 0.7746 & 1.0472 & \textbf{0.0042} & \textbf{0.0005} \\
& \textbf{LightGBM} & \color{darkgray}0.5282 & \color{darkgray}\textbf{0.5730} &  &  & 0.5282 & 0.5664 & 15.9230 & 0.8351 \\
& \textit{\textbf{EFG\textsubscript{ss}}} & \color{darkgray}\textit{0.4414} & \color{darkgray}\textit{0.5481} & \color{darkgray}\textit{0.4519} & \color{darkgray}\textit{0.5627} & \textit{0.4377} & \textit{0.5322} & \textit{547.1785} & \textit{86.1564} \\
& \textbf{EFG} & \color{darkgray}0.5084 & \color{darkgray}0.6010 & \color{darkgray}0.5209 & \color{darkgray}0.6211 & 0.5052 & 0.5855 & 122.0836 & 20.5225 \\
& \textbf{HOEG}& \color{darkgray}\textbf{0.4739} & \color{darkgray}0.5745 & \color{darkgray}\textbf{0.4836} & \color{darkgray}\textbf{0.5878} & \textbf{0.4700} & \textbf{0.5610} & 536.0755 & 68.5030 \\
\hdashline
\multirow{5}{*}{\rotatebox[origin=c]{90}{OTC}} & \textbf{Median} & \color{darkgray}0.7379 & \color{darkgray}0.9888 &  &  & 0.7175 & 0.8762 & \textbf{0.0064} & \textbf{0.0021} \\
& \textbf{LightGBM} & \color{darkgray}0.5422 & \color{darkgray}0.5021 &  &  & 0.6060 & \textbf{0.5980} & 1.9556 & 0.7211 \\
& \textit{\textbf{EFG\textsubscript{ss}}} & \color{darkgray}\textit{0.6359} & \color{darkgray}\textit{0.7769} & \color{darkgray}\textit{0.7338} & \color{darkgray}\textit{1.0440} & \textit{0.6585} & \textit{0.7560} & \textit{750.0486} & \textit{150.5996} \\
& \textbf{EFG} & \color{darkgray}\textbf{0.1835} & \color{darkgray}\textbf{0.1601} & \color{darkgray}\color{darkgray}\textbf{0.5985} & \color{darkgray}0.8779 & \textbf{0.5352} & 0.6951 & 83.8779 & 6.6364 \\
& \textbf{HOEG}& \color{darkgray}0.2163 & \color{darkgray}0.1804 & \color{darkgray}0.6069 & \color{darkgray}\textbf{0.8723} & 0.5505 & 0.6952 & 305.0114 & 25.1399 \\
\hdashline
\multirow{5}{*}{\rotatebox[origin=c]{90}{FI}} & \textbf{Median} & \color{darkgray}0.7673 & \color{darkgray}1.2763 &  &  & 0.7702 & 1.2881 & \textbf{0.0071} & \textbf{0.0005} \\
& \textbf{LightGBM} & \color{darkgray}0.7167 & \color{darkgray}\textbf{0.8035} &  &  & 0.7286 & \textbf{0.8334} & 15.8959 & 1.1631 \\
& \textit{\textbf{EFG\textsubscript{ss}}} & \color{darkgray}\textit{0.7250} & \color{darkgray}\textit{1.0537} & \color{darkgray}\textit{0.7222} & \color{darkgray}\textit{1.0338} & \textit{0.7310} & \textit{1.0579} & \textit{645.9569} & \textit{749.5316} \\
& \textbf{EFG} & \color{darkgray}0.6928 & \color{darkgray}0.9487 & \color{darkgray}\textbf{0.6906} & \color{darkgray}\textbf{0.9328} & \textbf{0.6955} & 0.9518 & 211.5477 & 21.3843 \\
& \textbf{HOEG} & \color{darkgray}\textbf{0.6879} & \color{darkgray}0.9911 & \color{darkgray}0.6919 & \color{darkgray}0.9898 & 0.6961 & 1.0037 &  884.2844 & 82.1531 \\
\hline
\end{tabular}%
}
\end{table}

Table \ref{tab:baseline_table} lists the results. We observe that Median achieves the worst performance and the best scalability. While the EFG\textsubscript{ss}-based GNN demonstrates superior performance in the case of BPI17, it ranks second to last among the models when applied to the other OCELs. LightGBM, then, strikes a balance between scalability and performance as it is the best scoring baseline on the OTC and FI OCELs.


Comparing this to EFG and HOEG, we notice that the EFG achieves the lowest score most frequently, meaning its predictions are closest to the actual values on average. Also, they do not deviate significantly from the true remaining times compared to the other models. 
%
However, when considering test MAE and MSE scores, this edge is modest to negligible on OTC and FI, and HOEG more strongly outperforms EFG on the BPI17 OCEL.

\section{Discussion} \label{sec:discussion}

\subsubsection{Performance and Scalability of HOEG}
The results concerning the performance, in terms of minimizing prediction errors, seems inconclusive. 
The results of experiments 2 and 3 show that HOEG performs significantly better than EFG and LightGBM for BPI17 in terms of test MAE and MSE, see \autoref{fig:exp2}, but slightly worse than EFG on the OTC and FI logs, as listed in~\autoref{tab:baseline_table}. One reason for these results would be the varying number of object features and event-object interactions in these three logs. 

In terms of scalability, HOEG, as anticipated, requires more training and prediction time compared to EFG, attributed to its additional hidden dimensions. However, the training duration of HOEG is comparable to, and its prediction time is shorter than, EFG with subgraph sampling (EFG\textsubscript{ss}), indicating its practical applicability.

\myparagraph{Object Attributes and Interactions}%
The three OCELs exhibit varying numbers of informative object variables and event-object interactions, see~\autoref{tab:ocels_summary}. BPI17 has 10 object features with 1.35 object interactions per event on average. OTC only has one (dummy) object feature, while FI presents a lower degree of object interaction on average. These differences in object interaction levels set the stage for understanding the machine learning model performance across these datasets.

\textit{BPI17} is relatively structured, having only two object types and originating from a WfMS. Each event is always related to one application, and only the latter events in a trace are related to one or more offers. Regarding Table~\ref{tab:baseline_table}, this may be why the graph-based models learn so well in this scenario. Furthermore, BPI17 has well-defined object attributes. These two characteristics seem to be leveraged by HOEG, as indicated by it having the lowest test MAE and MSE (when disregarding EFG\textsubscript{ss} for reasons discussed later).

\textit{OTC} has a high degree of object interaction, which was anticipated to favor HOEG's performance. Instead, we observe that EFG achieves a slightly lower MAE, and LightGBM outperforms HOEG in terms of MSE. Two factors may contribute to this unexpected outcome.
Firstly, the questionable validity of event attributes such as \texttt{weight} and \texttt{price} in the OTC dataset poses challenges. From an ontological standpoint, these attributes might be better suited as object attributes for specific object types (e.g., \textbf{order}, \textbf{item}, or \textbf{package}). The current inclusion of these attributes as event attributes implies a leakage of unavailable object information into events. That is, all events have values for \texttt{weight} and \texttt{price}, while some events might not relate to an object of the type that these attributes would belong to in a more realistic setting.
Secondly, OTC originally lacked object attributes. 
%
We might use this lack of object attributes and the questionable validity of event attributes to explain EFG slightly outperforming HOEG. That is, where EFG might leverage \texttt{weight} and \texttt{price}, HOEG has to filter out the noise of the object IDs as object attributes.

\textit{FI}, then, being extracted from an operational WfMS deployed in a highly complex process, has a lower level of both object interaction and process execution uniformity. HOEG seems to be surpassed by EFG by a negligible margin on MAE and was moderately outperformed by LightGBM in terms of test MSE. This might be explained by the process being extracted from a WfMS which is designed around a single case notion. This has led to different types of cases being forced into the same process, which might obscure the structuredness of process executions. This could result in a process that is difficult to predict. Furthermore, the vicinity of the MAE scores of Median and LightGBM might suggest a potentially limited predictive value of event and object attributes. That is, a naive predictor (Median), which does not learn from any attributes, approximates the performance of LightGBM which does learn complex patterns from attributes.

Therefore, HOEG is deemed promising \textit{for scenarios characterized by extensive object interaction.} 
Moreover, when paired with informative object attributes, the increased number of edges in HOEG may facilitate the faster passage of object attribute information through nodes. This potentially enhances the overall learning capacity of HOEG-based models.

\myparagraph{Subgraph Sampling} In Table~\ref{tab:baseline_table}, we observe that EFG\textsubscript{ss}, which uses subgraph sampling, seems to perform better than HOEG and EFG for BPI17, but worse for OTC and FI.  

We do note that this may not be a fair comparison. In essence, the subgraph sampling technique pools $k-1$ previous events to augment the current event in the encoding. 
%
Reimplementing the subgraph sampling in~\cite{adams2022-framework}, the technique excludes the first $k-1$ events per trace from the predictions. Therefore, the performance of the model on these first $k-1$ events is not included in the results. This implies that more samples are used for training and fewer samples are predicted.

Despite the potential performance improvement that may be achieved by using subgraph sampling (as listed in Table~\ref{tab:baseline_table}), its recommendation hinges on specific contextual factors. When scalability is a critical concern, subgraph sampling might not be advisable. That is, it requires the global pooling operation within the GNN architecture, which can be computationally expensive. 
%
%
Lastly, in high object interaction OCELs, subgraph sampling slices up relevant directly-follows edges, resulting in a series of unconnected events. For instance, in our illustrative Execution A (depicted in Figure \ref{fig:HOEG_concept}), taking a subgraph sample of size $4$ produces many samples of events that are unconnected (e.g., $<e1,e2,e3,e4>$).

\myparagraph{Design Choices and Trade-offs}
\ch{
%
The current HOEG approach assumes that the object features are immutable, meaning the features of each object type do not change over time, which may be an interesting direction to explore. 
Moreover, in our configuration of HOEG, object-event relations were given a generic \texttt{interacts} qualifier in $\mathcal{ET}$ and its adjacency matrices $\mathcal{A}$. Nevertheless, if a future log format enables descriptors for these interaction relations (e.g., (\texttt{order}, \texttt{createdby}, \texttt{event})), we can qualify these relations by configuring HOEG accordingly. This is supported already in our current definition of HOEG.
Finally, the current (\texttt{event}, \texttt{follows}, \texttt{event}) edge type and its adjacency matrix in HOEG simply reflect the edges between the events in a process execution. As shown, HOEG accommodates different extraction techniques. It therefore also supports future, more refined extraction techniques. 
}

\section{Conclusion} 
\label{sec:conclusion}
In the evolving field of PPM, this study sought to develop a general approach for encoding object-centric event data including their objects, interactions, and features. This approach facilitates the encoding without the need for flattening events, filling unavailable object features, or aggregating objects and their features.  Our efforts culminated in the creation of the HOEG approach.

Through three experiments performed on three datasets, we obtained insight into the efficacy of HOEG in terms of performance and scalability, particularly in predicting process remaining time. A key comparison was made with the EFG~\cite{adams2022-framework}. The results show that in scenarios involving well-structured object-centric processes, such as BPI17, HOEG-based GNNs perform better than EFG-based models. This result may suggest that HOEG is better suited for OCELs with abundant object interactions and well-defined attributes.

However, our study also revealed a trade-off in terms of scalability. HOEG-based models, with their more \emph{native} structure and increased parameter count, tend to be less scalable than EFG-based models. Despite this, considering both performance and scalability, HOEG stands out as a promising approach for leveraging \emph{native} data structures in OCELs, particularly for tasks like predicting the remaining time in processes.

Future work could explore various extensions of HOEG. Our study evaluated only one implementation, but others might include edge features or different heterogeneous GNN architectures. Additionally, the subgraph sampling technique might enhance model performance by more effectively capturing hierarchical patterns through pooling operations. 
Beyond our current focus, the HOEG approach offers flexibility to a variety of prediction tasks. Via HOEG one can train prediction models for different entities in the heterogeneous graph simultaneously, for instance, predicting for both events and objects, instead of creating a separate encoding as in~\cite{berti2022-graph_features_OCEL}. This flexibility opens up intriguing possibilities for applications like outlier detection and seems worthy to be explored.

\myparagraph{Acknowledgements} We thank Mike van Bussel for his invaluable contribution to this paper. 

%
%
%
\bibliographystyle{splncs04} 
\bibliography{references}

\begin{thebibliography}{10}
\providecommand{\url}[1]{\texttt{#1}}
\providecommand{\urlprefix}{URL }
\providecommand{\doi}[1]{https://doi.org/#1}

\bibitem{aalst2020-discovering_OC_PNs}
van~der Aalst, W.M.P., Berti, A.: Discovering object-centric petri nets. Fundam. Informaticae  \textbf{175}(1-4),  1--40 (2020)

\bibitem{adams2022-framework}
Adams, J.N., Park, G., Levich, S., Schuster, D., van~der Aalst, W.M.P.: A framework for extracting and encoding features from object-centric event data. In: {ICSOC}. Lecture Notes in Computer Science, vol. 13740, pp. 36--53. Springer (2022)

\bibitem{adams2022-cases_variants_oced}
Adams, J.N., Schuster, D., Schmitz, S., Schuh, G., van~der Aalst, W.M.P.: Defining cases and variants for object-centric event data. In: {ICPM}. pp. 128--135. {IEEE} (2022)

\bibitem{berti2022-graph_features_OCEL}
Berti, A., Herforth, J., Qafari, M.S., van~der Aalst, W.M.P.: Graph-based feature extraction on object-centric event logs. Int J Data Sci Anal  (2023)

\bibitem{bpic17}
van Dongen, B.F.: Bpi challenge 2017 (2017)

\bibitem{DBLP:books/sp/22/Fahland22}
Fahland, D.: Process mining over multiple behavioral dimensions with event knowledge graphs. In: Process Mining Handbook, Lecture Notes in Business Information Processing, vol.~448, pp. 274--319. Springer (2022)

\bibitem{galanti2023-ocppm}
Galanti, R., de~Leoni, M., Navarin, N., Marazzi, A.: Object-centric process predictive analytics. Expert Syst. Appl.  \textbf{213},  119--173 (2023)

\bibitem{ghahfarokhi2020-ocel}
Ghahfarokhi, A.F., Park, G., Berti, A., van~der Aalst, W.M.P.: {OCEL:} {A} standard for object-centric event logs. In: {ADBIS} (Short Papers). Communications in Computer and Information Science, vol.~1450, pp. 169--175. Springer (2021)

\bibitem{gherissi2023-ocppm}
Gherissi, W., Haddad, J.E., Grigori, D.: Object-centric predictive process monitoring. In: {ICSOC} Workshops. LNCS, vol. 13821, pp. 27--39. Springer (2022)

\bibitem{morris2019-HigherOrderGNN}
Morris, C., Ritzert, M., Fey, M., Hamilton, W.L., Lenssen, J.E., Rattan, G., Grohe, M.: Weisfeiler and leman go neural: Higher-order graph neural networks. {AAAI} pp. 4602--4609 (2019)

\bibitem{pourbafrani2022-rem_time_intercase}
Pourbafrani, M., Kar, S., Kaiser, S., van~der Aalst, W.M.P.: Remaining time prediction for processes with inter-case dynamics. In: {ICPM} Workshops. Lecture Notes in Business Information Processing, vol.~433, pp. 140--153. Springer (2021)

\bibitem{rohrer2022-thesis}
Rohrer, T., Ghahfarokhi, A.F., Behery, M., Lakemeyer, G., van~der Aalst, W.M.P.: Predictive object-centric process monitoring. CoRR  \textbf{abs/2207.10017} (2022)

\bibitem{smit2023}
Smit, T.K.: How object-centric is object-centric predictive process monitoring? (2023), \url{https://studenttheses.uu.nl/handle/20.500.12932/45369}

\bibitem{teinema2019-outcome_ppm_survey}
Teinemaa, I., Dumas, M., Rosa, M.L., Maggi, F.M.: Outcome-oriented predictive process monitoring: Review and benchmark. {ACM} Trans. Knowl. Discov. Data  \textbf{13}(2),  17:1--17:57 (2019)

\bibitem{you2020-design_space}
You, J., Ying, Z., Leskovec, J.: Design space for graph neural networks. In: NeurIPS (2020)

\end{thebibliography}
\end{document}